\documentclass[10pt]{article}
\usepackage{amsmath,mathrsfs,amssymb,amsthm,amscd}
\usepackage{graphicx}
\usepackage{geometry}
\usepackage{newpxtext}
\usepackage{appendix}
\usepackage{url}
\usepackage[table]{xcolor} 
\usepackage{subfig}
\usepackage[ruled,vlined]{algorithm2e}

\usepackage[utf8]{inputenc}

\usepackage{listings}
\usepackage{xcolor}

\definecolor{codegreen}{rgb}{0,0.6,0}
\definecolor{codegray}{rgb}{0.5,0.5,0.5}
\definecolor{codepurple}{rgb}{0.58,0,0.82}
\definecolor{backcolour}{rgb}{0.95,0.95,0.92}

\lstdefinestyle{mystyle}{
    backgroundcolor=\color{backcolour},   
    commentstyle=\color{codegreen},
    keywordstyle=\color{magenta},
    numberstyle=\tiny\color{codegray},
    stringstyle=\color{codepurple},
    basicstyle=\ttfamily\footnotesize,
    breakatwhitespace=false,         
    breaklines=true,                 
    captionpos=b,                    
    keepspaces=true,                 
    numbers=left,                    
    numbersep=5pt,                  
    showspaces=false,                
    showstringspaces=false,
    showtabs=false,                  
    tabsize=2
}

\theoremstyle{plain}
 \newtheorem{thm}{Theorem}[section]
 
 \numberwithin{equation}{section} 
 \theoremstyle{plain}
 \theoremstyle{plain}
 \theoremstyle{definition}
 \newtheorem{defn}[thm]{Definition}
 
 \theoremstyle{plain}
 \newtheorem{prop}[thm]{Proposition}
 
 \newtheorem{lem}[thm]{Lemma}
 
 \newtheorem*{cor*}{Corollary}
 \newtheorem*{conj*}{Conjecture}
 \newtheorem*{thm*}{Theorem}

\newcommand{\bl}{\begin{lem}}
\newcommand{\el}{\end{lem}}

\newcommand{\bml}{\begin{multline}}
\newcommand{\eml}{\end{multline}}

\newcommand{\beq}{\begin{equation}}
\newcommand{\eeq}{\end{equation}}
\newcommand{\bp}{\begin{prop}}
\newcommand{\ep}{\end{prop}}
\newcommand{\bd}{\begin{defn}}
\newcommand{\ed}{\end{defn}}
\newcommand{\pf}{\begin{proof}}
\newcommand{\epf}{\end{proof}}

\title{Edge Minimizing the Student Conflict Graph}
\author{J.S. Friedman\footnote{The views expressed in this article are the author's own and not those of the U.S. Merchant Marine Academy,
the Maritime Administration, the Department of Transportation, or the United States government.}}
\date{} 

\begin{document}

\maketitle

\begin{abstract}
In many schools, courses are given in sections. Prior to timetabling students need to be assigned to individual sections. We give a hybrid approximation sectioning algorithm that minimizes the number of edges (potential conflicts) in the student conflict graph (SCG). We start with a greedy algorithm to obtain a starting solution and then continue with a constraint programming based algorithm (CP-SAT) that reduces the number of edges. We apply the sectioning algorithm to a highly constrained timetabling model which we specify.
\end{abstract}

\section{Introduction} \label{secIntro}
Academic timetabling is the task of scheduling courses to specific times in such a way that there are no conflicts.  Most of the models considered in the literature assume that this conflict information is already known. However in many real life timetabling problems, courses are taught in multiple sections and until a student is assigned to a specific section of a course, the conflict information is not known.

M.W. Carter \cite{carter} sums it up nicely ``When courses are offered in multiple sections as they are at Waterloo, it creates a timetabling paradox. Students request a course, but timetabling assigns days and times to course sections. We cannot assign times to sections until we know which students are in each section. But we cannot assign students to sections until we know when the sections are timetabled!'' We call this paradox the \emph{student sectioning} problem.

One solution is to try to solve both sectioning and timetabling problems  simultaneously. When formulating 0-1 integer programs (IP) to model this situation, the resulting IP is huge and quickly become intractable: the search space is very large\footnote{According to our experiments, solving the combined IP with the Gurobi solver seems to take two orders of magnitude more time than separating out the sectioning first.}.

We briefly define two graphs of interest and give more details below. Suppose we have $n$ students and each has to enroll in a set of courses. Suppose each course can be split into different sections so when a student enrolls in a course they actually enroll in a specific section. Consider the \emph{course conflict graph} (CCG) where vertices are courses and edges signify at least one student must take the two courses incident with that edge. Next split each course of the CCG into sections. The \emph{student conflict graph} (SCG) is the graph where the vertices are the sections. A feasible student sectioning would induce edges in the SCG, where an edge connects two sections if a student is enrolled in both sections, or if a professor teaches the two sections, or if the two sections need the same room (specialty lab course). If one assigns students to sections in a random way the number of edges would be larger than necessary, and the chromatic number may be too high and/or it may be very difficult/impossible to find a feasible timetable.
 
In \cite{schindl} D. Schindl makes important progress on the student sectioning problem. In \cite{schindl} an \emph{optimal} sectioning of students is given in the special case where 1) all students have the same required list of courses, and 2) the courses are split into balanced sections. Schindl proves that a type of sectioning called \emph{regular sectioning} creates a minimal-edged SCG. Furthermore, the sectioning can be done quickly as it is not NP-hard. Though \cite{schindl} deals with a special case, it hinted to us that studying the SCG more systematically, investing more computational time upfront, could make timetabling much easier when sectioning is involved. Sectioning deserves much more attention in the literature.

In \cite{laporte} Laporte and Desroches considered the student sectioning problem in the situation where a complete timetable was already published. They first group students with identical schedules and merge them into a $k-student,$ and then try to optimally schedule them into sections, ignoring room capacities. Next they attempt to balance sections and respect room capacities, using local tree search techniques.

In \cite{carter} M.W. Carter also realized that the SCG is crucial to finding a feasible timetable, though he calls it the \emph{conflict matrix}. He defines a metric on the set of students that measures whether two students have similar course requests, and assigns students to sections so that two students who are near each other are likely to be assigned to the same sections. Once this assignment is done a timetable is computed, and then students are reassigned if necessary to different sections. Carter calls this sectioning \emph{homogeneous sectioning} and shows via experiment that it reduces the number of conflict edges by approximately twenty percent. 

In M{\"u}ller et al. \cite{murray,muller} they study the student sectioning problem at a large university. They start with an initial sectioning that based on Carter's homogeneous sectioning. Then a local search algorithm is used to improve on the initial sectioning using the following two moves:
\begin{itemize}
\item Two students enrolled in the same course swap all of their class assignments.
\item A student is re-enrolled into classes associated with a course such that the number of conflicts involving that student is minimized.
\end{itemize}
In \cite{murray,muller} they do not have all of the student request information at the time of timetable construction, so later on they do additional sectioning.

In \cite{selim} S.M. Selim approaches the student sectioning problem using a vertex splitting scheme. Before courses are split into sections one can form the course conflict graph (CCG). The chromatic number of this graph can exceed the number of available periods, and Selim gives careful conditions on how splitting course vertices into sections, so that the CCG becomes the SCG, can reduce the chromatic number of the final SCG. The real-life problem that Selim solves only has 12 different courses and each student takes three courses. It is not clear if his method would scale up to our model. 
 
The purpose of our paper is to generalize the work of \cite{schindl} on the student conflict graph (SCG) to the case where students all have a list of required courses, but they can differ from one student to the next; and the sections do not have to be balanced. The resulting problem is NP-hard\footnote{It is possible to relate our problem to a certain partitioning problem which is known to be NP-Hard. More details will follow in a future publication.} and hence we can not hope to find an optimal solution as Schindl does for his case. We give an approximation algorithm which produces a sectioning that gives an approximate minimal-edged SCG. We define the SCG to our more general case and we give a two-part algorithm that greatly reduces the number of edges (conflicts) in the SCG so that the corresponding (difficult) timetabling feasibility problem can be solved much more easily. The first part of the algorithm is a greedy approach similar in spirit to \cite{carter}, and second part utilizes the open source constraint programming solver called Google OR-Tools: CP-SAT \cite{ortools}.

\begin{figure}[h] 
\includegraphics[width=10cm]{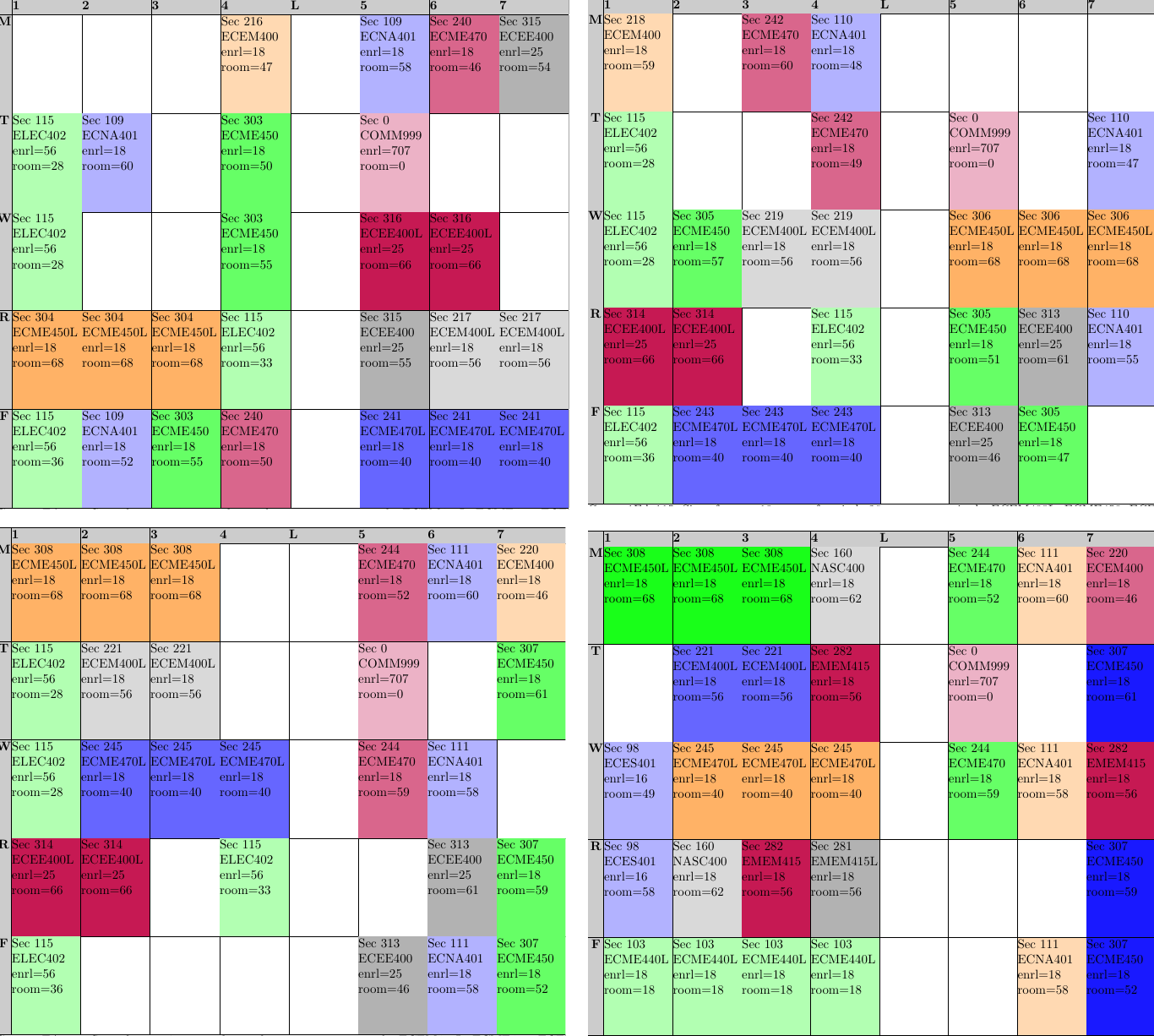}
\caption{\small In three of the schedules (1,2,3) the students have the same required courses but they are in mostly different sections. In the schedule (4)  students from a different major share some sections with (3). This allows for balanced sections and a minimal amount of teaching staff. } \label{fig:4sch}
\end{figure}

See Figure~\ref{fig:4sch} for an example of the type of timetabling problem we solve.  As one can see, our model allows a very dense student schedule and \emph{all} courses are required. This model is applicable to any school that wants to maximize student learning in a minimal amount of time. In fact, this is the model used at the U.S. Merchant Marine Academy where students essentially earn three degrees in four years, spending one of those years at sea. They receive an officer commission in the U.S. Navy, a U.S. Coast Guard license, and a rigorous B.S. degree, all while earning upwards of 160 credits in certain majors. Clearly this model is not for very large universities, but for schools who wish to become more efficient it might be advantageous.  

Ultimately the purpose of student sectioning is to construct a timetable. Our model is not usually studied in the literature. Though small compared to large university models, our is meant for less students but with really packed schedules. We split the week into equal sized periods, for example seven one-hour periods per day, 5 days per week. The model includes lectures (3 or 4 periods a week, one meeting per day) and extended lectures (multiple periods in a row). Taking one look at Figure~\ref{fig:1mc2sc} shows that predefined set course times would extend the day in our model, as our model can pack in more courses more efficiently. What makes our model challenging are the numerous extended lectures and the dense amount of periods used. In the timetabling process it is very difficult to move an extended section as they \emph{anchor} the timetable and conflict with other sections that do not have the same footprint. The problem is highly constrained (sometimes over-constrained). The model that we propose has the potential to utilize time and space more efficiently, especially if the majors have significant overlap and there are many required courses. Our model is student centric, and students are \emph{guaranteed} a spot in all of their courses.

We have created instances of various difficulty of our models and posted the source code for our work, as well as a solved timetable on \url{(https://github.com/mathprofessor/timetable)}.

\section{Model}
\subsection{Preliminaries}
We consider the following model for student sectioning and timetabling. Let $G$ denote the set of \emph{students.} We partition $G$ into \emph{major-groups} $M$ where $\bigcup_{m \in M} m = G.$ Let $C$ denote the set of \emph{courses}, and $S$ the set of \emph{sections.} For each course $c \in C$ we have a set of sections $S_c,$ where the $S_c$ partition $S.$ For example, Calc1 would represent a course, and Calc1.01, Calc1.02,... would be the corresponding sections. When a student enrolls in a course they select exactly one section. Each student $g$ in a fixed major-group $m$ is required to enroll in a set of \emph{courses} $C_m.$ More precisely, each student must enroll in a particular section of each course in $C_m.$ Some sections are tied together in the parent/child relationship like lectures and labs. Section $s_1$ is the unique \emph{parent} of $s_2$ if enrollment in $s_2$ implies enrollment in $s_1;$ conversely $s_2$ is a \emph{child} of $s_1,$ and the set of sections consisting of a parent and all children is a \emph{family}. Our model precludes grandchildren and to limit complexity of the model we insist that all members of a family have the same capacity.\footnote{In real-life models one would need a \emph{satillite-family} where a big lecture branches into smaller recitations and labs and the capacities of the members differ. Our model can easily be extended to include this.} Note that we consider parent and child (and their corresponding courses) as separate entities as they generally have different staffing and room requirements.

\begin{figure}[h] 
    \centering
    \subfloat[\centering \tiny Division 1MC.1 \mbox{7 students} ]{{\includegraphics[width=6cm]{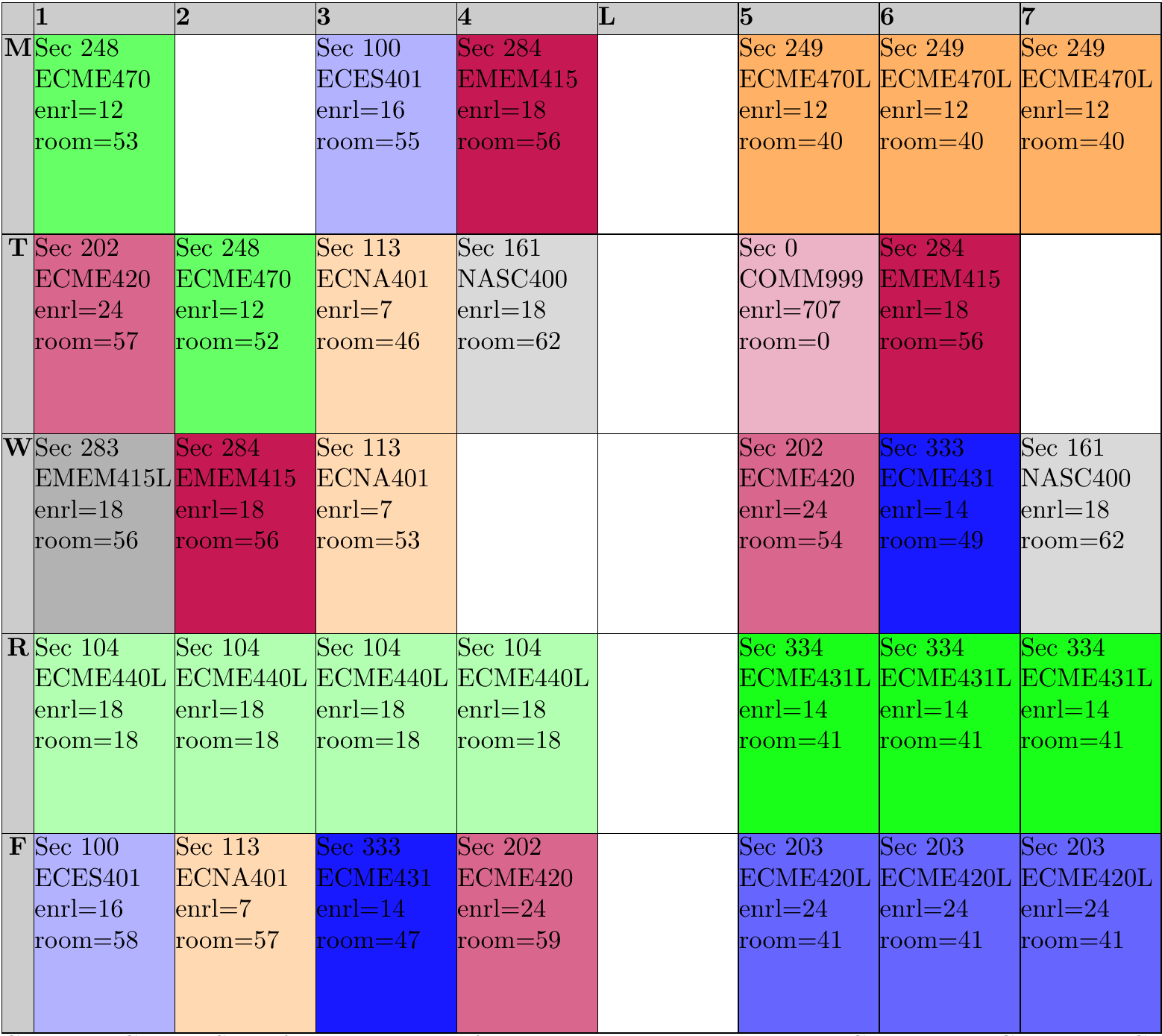} }}
    \qquad
    \subfloat[\centering \tiny Division 2SC.2 \mbox{9 students} ]{{\includegraphics[width=6cm]{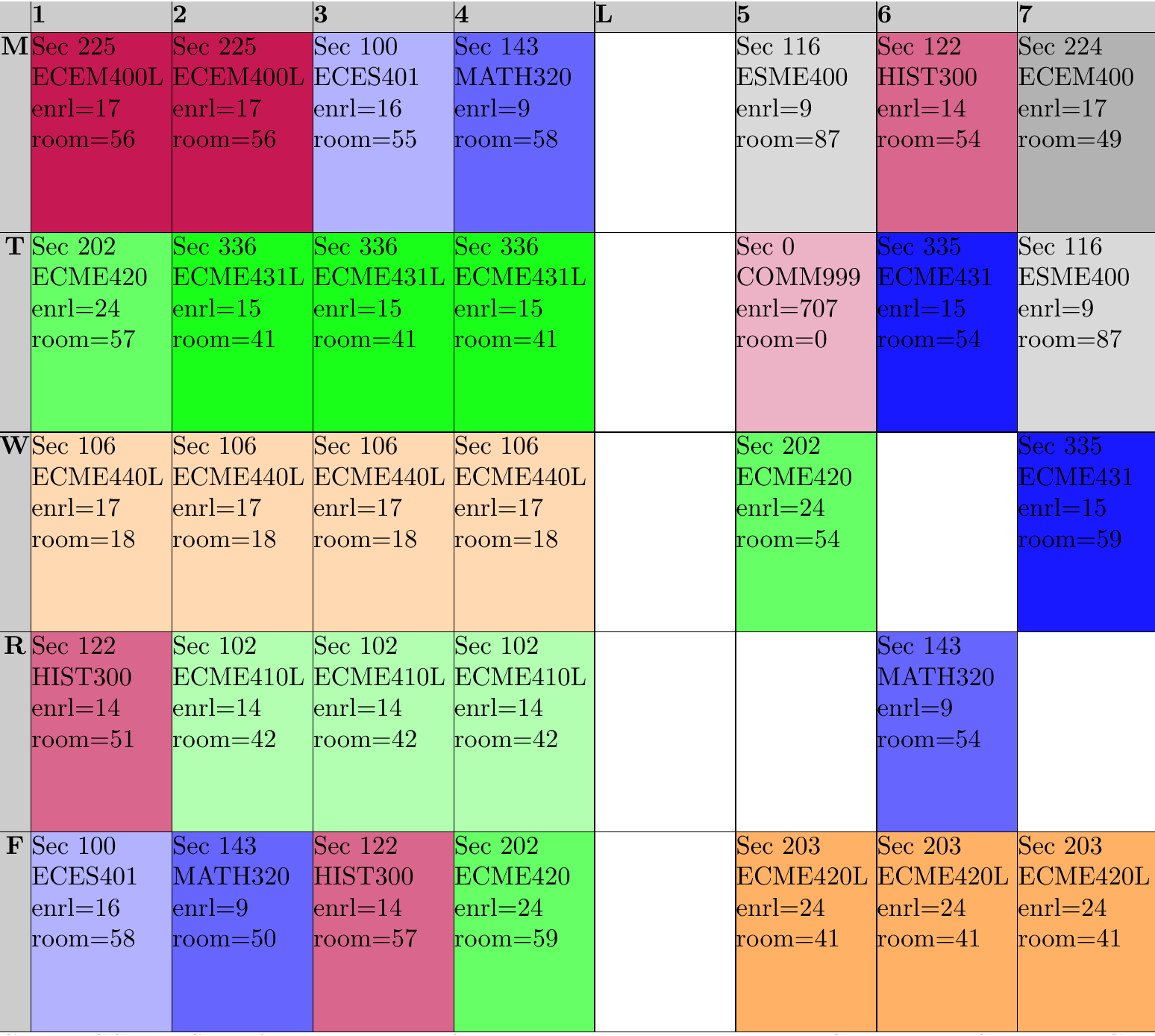} }}
    \caption{\small Note how the two divisions share sections 100, 202, 203 even though they require a different set of courses. The two divisions come from two different major-groups 1MC and 2SC. }
    \label{fig:1mc2sc}
\end{figure}

\begin{figure}[h]
    \centering
    \subfloat[\centering \tiny Division 4EX.1 \mbox{3 students} ]{{\includegraphics[width=6cm]{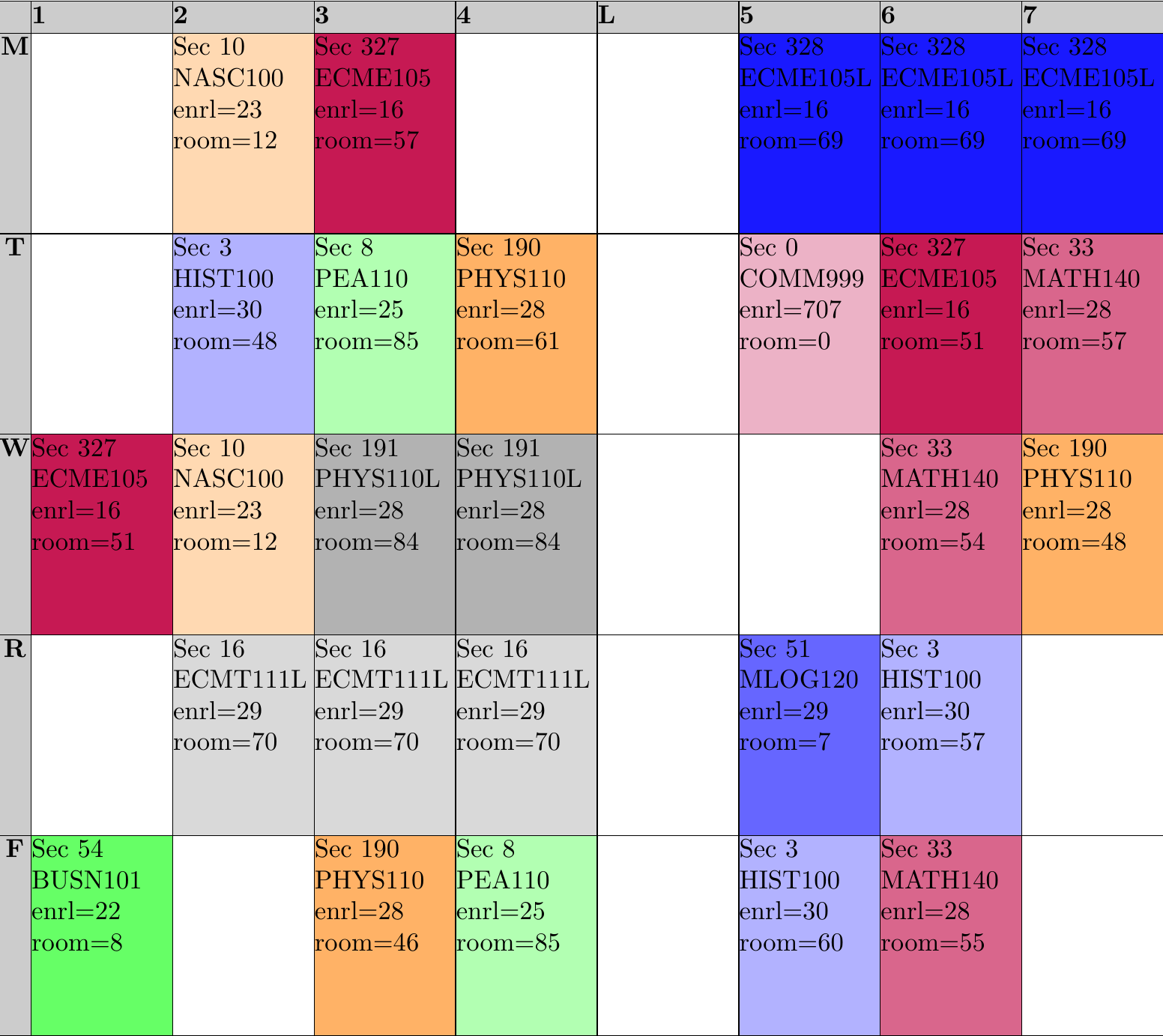} }}
    \qquad
    \subfloat[\centering \tiny Division 4EX.2 \mbox{11 students} ]{{\includegraphics[width=6cm]{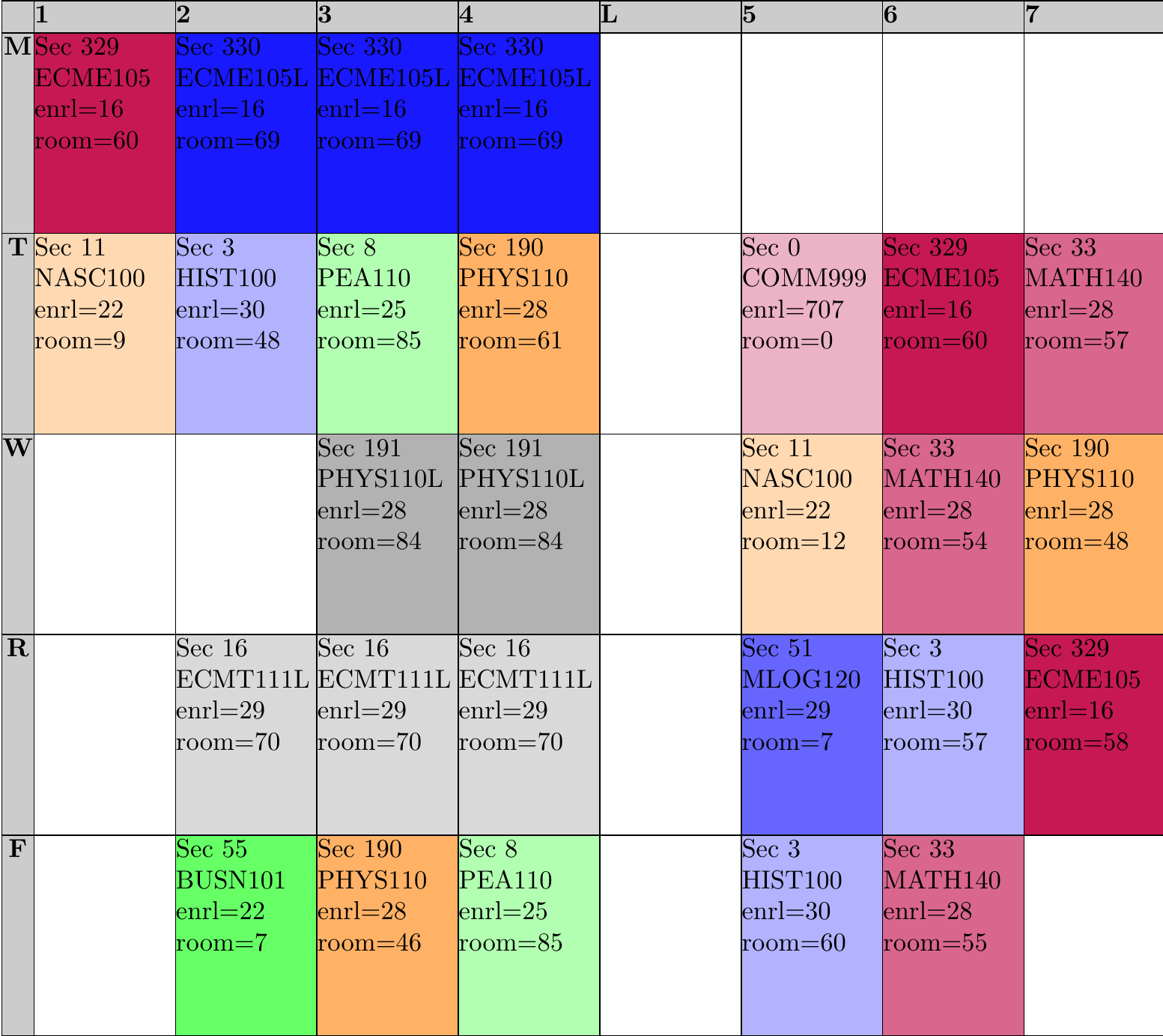} }}
    \caption{\small Note how the two divisions are in the same major-group 4EX (they require the same courses) but some of their sections differ. }
    \label{fig:4ex}
\end{figure}

The major-group is not a standard concept so we illustrate with an example. consider a small college or a large high school that offers 5 majors: Arts, Math, Engineering, Computers, and Biology. Suppose there are 4 class years: Freshman, Sophomore, Junior, Senior, and each class year is split into Honors or Standard. This would correspond to $(5)(4)(2) = 40$ major-groups. Each student in a major-group has identical course requirements but they can  enroll in different sections. There would be courses common to distinct major-groups (core courses that everyone takes). In addition, there could be major-groups of students who failed courses and must retake them and advanced students with unique course requirements. So in practice, there could be more than 40 major-groups, and it is possible that a major-group could include a single student. In our model elective choices are not applicable: any student can enroll in a section of any course, in theory they can have a unique schedule.
 
We partition students $G$ into \emph{divisions}, $D,$ where $\bigcup_{d \in D} d = G,$  where each student in a division $d$ has an identical section schedule. It follows that each division is subset of a unique major-group. See Figure~\ref{fig:1mc2sc} and Figure~\ref{fig:4ex} for more illustration.

Let $R$ denote the set of \emph{rooms.} We partition $R$ into \emph{room-types} $\text{RT}$ where 
$\bigcup_{q \in RT} q = R.$ Let $P$ denote the set of professors. For each section $s \in S$ we assign (before any sectioning) a professor $p=p(s)$ and a room-type $q = q(s).$  

Suppose that each major-group $m$ has size $|m|$ and each section $s$ has capacity  $|s|.$ The problem we consider is how to split the required courses into sections so each student of each major-group can enroll in a course without violating the capacity of the sections. 

\subsection{Student Conflict Graph}
For each student $g \in G,$ there is a unique major-group $m_g$ with $g \in m_g.$ Let $C_g = C_m$ denote the required courses of $g.$ Define the set $\text{GC} = \{ (g,c) ~|~ g \in G, c \in C_g\}.$ For each section $s,$ $\pi(s)$ is the corresponding course associated to $s,$ that is $s \in S_{\pi(s)}.$ 
Define the set $W = \{ (g,s) ~|~ g \in G, s \in \pi^{-1}(C_g) \}.$  In other words, $(g,s) \in W$ if $s$ is a section that student $g$ could enroll in. 

\begin{defn} \label{defSec}
We define a \emph{sectioning} to be a function $f:\text{GC} \mapsto W$ so the following properties are satisfied: 

\begin{enumerate}
\item If $f(g,c) = (g,s) $ then $s \in S_c.$ In this case we say $g$ is \emph{assigned} to section $s.$
\item For each $s \in S,$ the number of students assigned to $s$ does not exceed $|s|.$
\item If $s_1$ is the parent of $s_2$ then $f(g,\pi(s_2)) = (g,s_2)$ implies $f(g,\pi(s_1)) = (g,s_1).$

\end{enumerate}

\end{defn}

\begin{defn} \label{defSCG} Let $f$ be a sectioning. The student conflict graph of $f,$ $\text{SCG}(f)$ is the graph whose vertices are the sections $s \in S,$ and whose edges are given by  
\begin{enumerate}
\item $(s_1,s_2)$ if there is a professor who teaches both sections $s_1$ and $s_2;$
\item $(s_1,s_2)$ if $s_1$ and $s_2$ have the same room-type $q(s_1)=q(s_2),$ and the number of available rooms in the room-type $q(s_1)$ is exactly one;
\item $(s_1,s_2)$ if there is a student $g$ who is assigned to both sections $s_1$ and $s_2$ with respect to the sectioning $f.$ 
\end{enumerate}
\end{defn}

\section{Minimizing Edge Algorithms}
Our algorithm is a hybrid of two parts. The first is a greedy algorithm similar in spirit to \cite{carter}, which will give an initial solution far superior to randomly assigning students to sections. The results of the first part are passed to an open source constraint solver (CP-SAT) that further improves the objective. 

\subsection{Algorithm: Greedy Sectioning}
Let $G$ be the set of students. We define a distance function on $G$ by 
$$d(g_1,g_2) = \left|C_{g_1} \ominus C_{g_2}\right|, \quad \text{where $\ominus$ is the symmetric set difference.}$$

\begin{algorithm}[H]
\SetAlgoLined
\KwResult{Heuristic algorithm to create an approximate edge-minimal SCG. }

\textbf{Initialization:} Create the base SGC: For each $s \in S$ add a vertex. Add all edges $(s_1,s_2)$ whenever the first two items of Definition~\ref{defSCG} apply.
 
Choose a random student $g_0 \in G$ and randomly assign $g_0$ to sections, $S_{g_0},$ consistent with Definition~\ref{defSec}. Add edges to the SGC consistent with third item of Definition~\ref{defSCG}.

assign $g := g_0$
 \;
 \While{While there are still students who need to enroll}{

Choose an unenrolled student $h$ so $d(h,g)$ is minimal. 

Let $S_{gh}$ be the open sections of $S_g$ that $h$ also needs to enroll in; 

Enroll $h$ in all courses of $S_{gh};$
 
For each course $c$ in $C_h$ that is still needed, choose the section $s$ of $c$ so that it will add the minimal amount of new edges to the SGC, remembering the sections $h$ is enrolled in so far from $g$. Add the rest of the family corresponding to $s$ if needed. 

Assign $g := h$
 \;
 }
 \caption{Greedy Sectioning}
\end{algorithm}

The initial algorithm Greedy Sectioning, above, attempts to assign students to sections by ordering the students in a chain and copying the schedule, as much as possible, of previously scheduled students who have the most similar schedule. 

Greedy Sectioning has a runtime that is polynomial in the number of students and sections, and is quite fast.

\subsection{Algorithm: CP-SAT Sectioning}
Our algorithm is very similar to an 0-1 integer program. We model the algorithm using standard constraints from integer and constraint program. OR-Tools CP-SAT converts it to a SAT instance which is solved by the CP-SAT solver \cite{ortools}. We import the starting solution from Greedy Sectioning as a warm start (more specifically OR-Tools calls it a hint for the value of each variable).

There will be three options for the objective of the constraint program. The first tries to minimize the number of edges; the second puts a weight on edges associated with sections who use contiguous periods (since they are difficult to move in the construction of a timetable); and the third incorporates a signal from a failed attempt to timetable by adding some weight for \emph{tabu} assignments that the solver should stay clear of. Initially the third option is unavailable. We have included the Python source code in Appendix~\ref{appSec}.

We define some needed sets to formulate the CP. Let 
\begin{enumerate}
\item $W = \{ (g,s) ~|~ g \in G, s \in \pi^{-1}(C_g) \}.$
\item $\text{GC} = \{ (g,c) ~|~ g \in G, c \in C_g\}.$
\item $\text{SS} = \{(s_1,s_2) \in S \times S ~|~ s_1 < s_2 \}.$
\item $\text{GSS} = \{(g,s_1,s_2) \in G\times \text{SS} ~|~ (g,s_1),(g,s_2) \in W  \}.$
\item $\text{PS} = \{(p,s) \in P\times S ~|~ \text{$p$ teaches section $s$}\} $
\item $\text{PSS} = \{(p,s_1,s_2) \in P\times \text{SS} ~|~ (p,s_1), (p,s_2) \in \text{PS} \} $
\item $\text{FSS} = \{(s_1,s_2) \in S \times S ~|~ s_1 \neq s_2, \,\, \text{$s_1$ is the parent of $s_2$} \}.$
\item $\text{FGSS} = \{(g,s_1,s_2) \in G \times S \times S ~|~ (s_1,s_2) \in \text{FSS}, (g,s_1) \in W\}.$
\item $\text{RSS} = \{(r,s_1,s_2) \in R \times \text{SS} ~|~ \text{$s_1$ and $s_2$ require the exact same room} \} $
\end{enumerate}

Next we define \textbf{boolean} variables
\begin{enumerate}
\item $x_{g,s}$ for $(g,s) \in W.$ Here $x_{g,s} = 1$ iff student $g$ is enrolled in section $s.$
\item $y_{s,t}$ for $(s,t) \in \text{SS}.$ Here $y_{s,t} = 1$ iff sections $s,t$ can not be scheduled at the same time.
\end{enumerate}

The constraints are given by 
\begin{enumerate}
\item for $(s,t) \in \text{RSS} \cup \text{PSS},\quad $  $y_{s,t} = 1,$ (both sections have the same professor or need the same room)
\item for $(g,c) \in \text{GC}, \quad $ $\sum_{s \in S_c} x_{g,s} = 1,$ (choose one section from course $c$)
\item for $s \in S, \quad $ $$\sum_{\substack{g \in G \\ (g,s) \in W}} x_{g,s} \leq |s|,$$ (do not exceed capacity of section $s$)
\item for $(g,s_1,s_2) \in \text{GSS}, \quad $ $x_{g,s_2} \leq x_{g,s_1}, $ (if $s_2$ has parent $s_1$ then if we enroll in $s_2$ we must enroll in $s_1$)
\item for $(g,s_1,s_2) \in \text{GSS}, \quad $  $\text{Boolean\_OR}(\text{not}(x_{g,s_1}),\text{not}(x_{g,s_2}),y_{s_1,s_2} ).$ This constraint could also be written as $x_{g,s_1} + x_{g,s_2} - 1 \leq y_{s_1,s_2} $ if one is using a MIP solver, (if student $g$ is enrolled in sections $s_1,s_2$ then $y_{s_1,s_2}$ must be one.
\end{enumerate}

We have three choices for the objective. We minimize 
\begin{enumerate}
\item $Z = \sum_{(s,t) \in \text{SS}} y_{s,t},$ here we are minimizing the number of edges in the SCG. Minimizing this objective makes it easier to find a feasible coloring of the SCG which leads to a feasible timetable. However, in our model we have extended sections which take up 2,3, or 4 contiguous periods. The next objective takes this into account.
\item For $(s,t) \in \text{SS}$ define a weight $w_{s,t}$ by
$$w_{s,t} = \begin{cases} 
      a & \text{ if $s$ and $t$ are both not extended sections } \\
      b & \text{ if exactly one of $s$ and $t$ are extended sections } \\
      c & \text{ if $s$ and $t$ are both extended sections } 
   \end{cases} $$
where $a,b,c$ are appropriately chosen real numbers. Thus $Z = \sum_{(s,t) \in \text{SS}} w_{s,t} y_{s,t},$

   The rational for this is that extended sections are much harder to move in the construction of a timetable, and the fewer edges incident with extended sections, the better.
\item Sometimes after minimizing the edges of the SCG the timetabling process returns with sections that are in conflict.  See Appendix~\ref{appTime} lines 195--204 for the details in Python. These conflicts stem from unfortunate assignments of particular students to particular sections. We place these assignments in a tabu list and send it back to the CP-SAT Sectioning algorithm and restart the edge minimization where it left off but with a new objective:
$$Z = \sum_{(s,t) \in \text{SS}} w_{s,t} y_{s,t} + d\sum_{(g,s) \in \text{tabu}} x_{g,s},  $$
where $d > 0$ is chosen appropriately. This modification will discourage the tabu assignments, and starting the search with a warm start (previous solution, which is still feasible) would result in different sectioning.
\end{enumerate}

\section{Timetabling}
Once we have an attractive sectioning, or even better an entire collection of them computed in parallel, we solve our timetabling model. Each section $s$ requires a number of periods $\text{per}(s)$ and it is either \emph{extended} or not. Extended sections have all meetings contiguous and can not be separated by the lunch period. Our model has one common section that all students must take. Non-extended sections should have no more than one meeting per day. Sections that are adjacent on the SCG should not be scheduled at the same time, and professors should have one day where they do not teach\footnote{In our experiments we randomly requested a certain day off.}. Room capacity constraints are implicitly defined by specifying the room-type of each section.\footnote{In a real-life timetabling application a more specific approach to rooms is required, but our model is a good approximation.}

Our implementation is a boolean (0-1) constraint program (CP) using Google's OR-Tools CP-SAT solver, Python interface. Our CP utilizes standard constructs and the interested reader can refer to Appendix~\ref{appTime} where the Python code can be read quite easily. In addition the source code is available on \url{https://github.com/mathprofessor/timetable}. 

Our strategy is to first solve the sectioning problem and construct the SCG. Next each constraint is implemented as a soft constraint with a certain weight in the linear objective. For example, a section clash has weight 1000, while running out of rooms in a certain room-type has weight 100. A professor not getting a day off from teaching has less weight as does scheduling two or more meetings of a non-extended class on the same day. The reason for the soft constraints is our problems can be over constrained and the CP should indicate if and how we need to change our instance to obtain feasibility. Our goal here is a feasible timetable\footnote{Once a feasible timetable exists one would add other soft constraints and either re-solve the global problem or performing local searches by slicing up the periods and allowing certain groups of periods to move while keeping others fixed.}.

We first attempt to timetable only the extended sections and the common section. Once a feasible solution is found (usually optimal with respect to our soft constraints) we pass this solution to the CP and attempt to timetable the entire problem.

If we can not find a feasible timetable, we might only need a different SCG, and we can send the tabu data back to the sectioning program and try again, or try a different SCG. Typically we can solve these difficult problems in 5--30 minutes. See \S\ref{secBench} for more details on our experiments.

\section{Computational setup and benchmarks} \label{secBench}
Our computational machine has a AMD RYZEN 7 2700X 3.7GHZ CPU with 8 cores, capable of 16 threads, with 16GB of ram and 16GB or swap space\footnote{The CP-SAT needs a lot of memory.}. We implemented our code using Python and used Google's OR-Tools CP-SAT \cite{ortools}. We invoked the solver with 16 threads, and unlike MIP solvers, we experienced great speedup by using more threads. We also experimented with the Gurobi solver, and we found that the free open source CP-SAT solver is more efficient for these type of problems: sectioning and timetabling problems of this kind do not have a tight relaxation, and MIP solvers may not be the best choice. 

Four instances have been created (see \url{https://github.com/mathprofessor/timetable}), Easy, Medium, Medium2, and Hard, with 256, 339, 352, and 372  distinct sections respectively.

For each instance we ran the sectioning algorithm for (100,600,1800) seconds, three times, and then allocated 600 seconds to the timetable CP. Looking at the data below, when running the CP-SAT edge reduction, we can remove roughly 7\% of the edges after the greedy algorithm, and looking at the last three lines, makes it possible to solve the hard instance. 

\begin{table}[htbp] 
\tiny
\begin{tabular}{|l|l|l|l|l|l|l|}
\hline
\textbf{\begin{tabular}[c]{@{}l@{}}CP-SAT\\ seconds\end{tabular}} & \textbf{Instance} & \textbf{\begin{tabular}[c]{@{}l@{}}Size of SCG \\ after Greedy\end{tabular}} & \textbf{\begin{tabular}[c]{@{}l@{}}Size of SCG \\ after CP-SAT\end{tabular}} & \textbf{\begin{tabular}[c]{@{}l@{}}Percent reduction\\  of edges\end{tabular}} & \textbf{Objective} & \textbf{\begin{tabular}[c]{@{}l@{}}Time to optimal \\ solution of zero\end{tabular}} \\ \hline
\textbf{100}                                                      & easy              & 2616                                                                         & 2531                                                                         & 3.25                                                                           & 0                  & 38                                                                                   \\ \hline
\textbf{100}                                                      & easy              & 2619                                                                         & 2561                                                                         & 2.21                                                                           & 0                  & 36                                                                                   \\ \hline
\textbf{100}                                                      & easy              & 2619                                                                         & 2572                                                                         & 1.79                                                                           & 0                  & 37                                                                                   \\ \hline
\textbf{100}                                                      & medium            & 3992                                                                         & 3849                                                                         & 3.58                                                                           & 3                  &                                                                                      \\ \hline
\textbf{100}                                                      & medium            & 3976                                                                         & 3825                                                                         & 3.80                                                                           & 14                 &                                                                                      \\ \hline
\textbf{100}                                                      & medium            & 3970                                                                         & 3785                                                                         & 4.66                                                                           & 18                 &                                                                                      \\ \hline
\textbf{100}                                                      & medium2           & 4343                                                                         & 4199                                                                         & 3.32                                                                           & 13                 &                                                                                      \\ \hline
\textbf{100}                                                      & medium2           & 4319                                                                         & 4183                                                                         & 3.15                                                                           & 0                  & 148                                                                                  \\ \hline
\textbf{100}                                                      & medium2           & 4355                                                                         & 4194                                                                         & 3.70                                                                           & 16                 &                                                                                      \\ \hline
\textbf{100}                                                      & hard              & 4885                                                                         & 4683                                                                         & 4.14                                                                           & 238                &                                                                                      \\ \hline
\textbf{100}                                                      & hard              & 4825                                                                         & 4689                                                                         & 2.82                                                                           & 32                 &                                                                                      \\ \hline
\textbf{100}                                                      & hard              & 4830                                                                         & 4663                                                                         & 3.46                                                                           & 53                 &                                                                                      \\ \hline
\textbf{600}                                                      & easy              & 2619                                                                         & 2541                                                                         & 2.98                                                                           & 0                  & 33                                                                                   \\ \hline
\textbf{600}                                                      & easy              & 2612                                                                         & 2512                                                                         & 3.83                                                                           & 0                  & 37                                                                                   \\ \hline
\textbf{600}                                                      & easy              & 2612                                                                         & 2505                                                                         & 4.10                                                                           & 0                  & 37                                                                                   \\ \hline
\textbf{600}                                                      & medium            & 3993                                                                         & 3771                                                                         & 5.56                                                                           & 1                  &                                                                                      \\ \hline
\textbf{600}                                                      & medium            & 4036                                                                         & 3667                                                                         & 9.14                                                                           & 2018               &                                                                                      \\ \hline
\textbf{600}                                                      & medium            & 3987                                                                         & 3666                                                                         & 8.05                                                                           & 0                  & 90                                                                                   \\ \hline
\textbf{600}                                                      & medium2           & 4339                                                                         & 4090                                                                         & 5.74                                                                           & 2                  &                                                                                      \\ \hline
\textbf{600}                                                      & medium2           & 4343                                                                         & 4132                                                                         & 4.86                                                                           & 13                 &                                                                                      \\ \hline
\textbf{600}                                                      & medium2           & 4390                                                                         & 4114                                                                         & 6.29                                                                           & 5                  &                                                                                      \\ \hline
\textbf{600}                                                      & hard              & 4827                                                                         & 4583                                                                         & 5.05                                                                           & 52                 &                                                                                      \\ \hline
\textbf{600}                                                      & hard              & 4831                                                                         & 4596                                                                         & 4.86                                                                           & 8                  &                                                                                      \\ \hline
\textbf{600}                                                      & hard              & 4815                                                                         & 4565                                                                         & 5.19                                                                           & 10536              &                                                                                      \\ \hline
\textbf{1800}                                                     & easy              & 2633                                                                         & 2495                                                                         & 5.24                                                                           & 0                  & 35                                                                                   \\ \hline
\textbf{1800}                                                     & easy              & 2619                                                                         & 2501                                                                         & 4.51                                                                           & 0                  & 32                                                                                   \\ \hline
\textbf{1800}                                                     & easy              & 2612                                                                         & 2495                                                                         & 4.48                                                                           & 0                  & 35                                                                                   \\ \hline
\textbf{1800}                                                     & medium            & 3970                                                                         & 3650                                                                         & 8.06                                                                           & 0                  & 119                                                                                  \\ \hline
\textbf{1800}                                                     & medium            & 3992                                                                         & 3647                                                                         & 8.64                                                                           & 24                 &                                                                                      \\ \hline
\textbf{1800}                                                     & medium            & 3997                                                                         & 3653                                                                         & 8.61                                                                           & 0                  & 82                                                                                   \\ \hline
\textbf{1800}                                                     & medium2           & 4356                                                                         & 4119                                                                         & 5.44                                                                           & 7                  &                                                                                      \\ \hline
\textbf{1800}                                                     & medium2           & 4347                                                                         & 4061                                                                         & 6.58                                                                           & 0                  & 130                                                                                  \\ \hline
\textbf{1800}                                                     & medium2           & 4342                                                                         & 4064                                                                         & 6.40                                                                           & 0                  & 167                                                                                  \\ \hline
\textbf{1800}                                                     & hard              & 4847                                                                         & 4530                                                                         & 6.54                                                                           & 22                 &                                                                                      \\ \hline
\textbf{1800}                                                     & hard              & 4844                                                                         & 4482                                                                         & 7.47                                                                           & 15                 &                                                                                      \\ \hline
\textbf{1800}                                                     & hard              & 4844                                                                         & 4461                                                                         & 7.91                                                                           & 0                  & 199                                                                                  \\ \hline
\end{tabular}
\end{table}


\begin{appendices}

\section{Sectioning CP-SAT Program} \label{appSec}

\begin{tiny}
\begin{lstlisting}[language=Python]
def secCP(threads, flist, time,x_m = None,y_m = None, tabu1 = None):
	print('CP-SAT Sectionize')
	(S,R,RT,C,P,G,L,D,T,Lunch,sd,rd,pd,gd,catalog,divSizeDict,divDict) = flist
	W=set([(g,s) for g in G for s in S if (sd[s].course_num in gd[g].course_num_list)])
	GC = [(g,c) for g in G for c in C if c in gd[g].course_num_list ]	
	SS = [(s1,s2) for s1 in S for s2 in S if s1 < s2]
	GSS = [(g,s1,s2) for g in G for (s1,s2) in SS if ( (g,s1) in W and (g,s2) in W )  ]
	PS = [(p,s) for p in P for s in S if s in pd[p].secNums]
	PSS = [(p,s1,s2) for p in P for (s1,s2) in SS if ( (p,s1) in PS and (p,s2) in PS )  ]
	labtieSS = [(s1,s2) for s1 in S for s2 in S if sd[s1].labtie != '' and sd[s1].labtie == sd[s2].labtie and sd[s1].iAmParent and s1 != s2 ]
	GlabtieSS = [(g,s1,s2) for g in G for (s1,s2) in labtieSS if (g,s1) in W
	
	model = cp_model.CpModel()
	x = {(g,s) : model.NewBoolVar('x%i%i' % (g,s) ) for (g,s) in W}

	y = {}
			
	for (s,t) in SS:
		y[s,t] = model.NewBoolVar('x%i%i' % (s,t) )

	
	weight1 = {}
	for (s1,s2) in SS:
		wei = 1
		if sd[s1].isExtended:
			wei+=3
		if sd[s2].isExtended:
			wei+=3	
		weight1[s1,s2] = wei
	
	if tabu1 != None:
		model.Minimize( sum(weight1[s,t]*y[s,t] for (s,t) in SS ) + sum(5*x[g,s] for (g,s) in tabu1))
	else:
		model.Minimize( sum(weight1[s,t]*y[s,t] for (s,t) in SS ) )
	
	if x_m != None:
		for (g,s) in W:
			if (g,s) in x_m.keys():
				if x_m[g,s] == 1:
					model.AddHint(x[g,s],1)
			
		for (g,s) in W:
			if (g,s) not in x_m.keys():	
				model.AddHint(x[g,s],0)
		
		for (s1,s2) in SS:
			model.AddHint(y[s1,s2], y_m[s1,s2])
	
	rtCount = {}
	for r in R:
		if rd[r].roomtype in rtCount:
			rtCount[rd[r].roomtype]+=1
		else:
			rtCount[rd[r].roomtype] = 1

	SoneRoom = [s for s in S if rtCount[sd[s].roomtype] == 1 ]

	
	RSS = [(r,s1,s2) for r in R for s1 in SoneRoom for s2 in SoneRoom if s1 < s2 if sd[s1].roomtype == sd[s2].roomtype]
	
	for (r,s1,s2) in RSS:
		model.Add( y[s1,s2] == 1 )
	
	for (p,s1,s2) in PSS:
		model.Add( y[s1,s2] == 1 )
		
	for (g,c) in GC:
		model.Add(sum(x[g,s] for s in S if sd[s].course_num == c ) == 1  )
 	
	for s in S:
		model.Add(sum(x[g,s] for g in G if (g,s) in W ) <= sd[s].cap)	

	for (g,s1,s2) in GlabtieSS:	
		model.Add(x[g,s2] <= x[g,s1])
		
	for (g,s1,s2) in GSS:
		model.AddBoolOr([x[g,s1].Not(), x[g,s2].Not(), y[s1,s2]])		

	print('Ready to Solve')
	solver = cp_model.CpSolver()
	print('Created Solver')
#	solver.parameters.log_search_progress = True

	solver.parameters.search_branching = (cp_model.sat_parameters_pb2.SatParameters.PORTFOLIO_WITH_QUICK_RESTART_SEARCH)	
	solver.parameters.linearization_level = 0
	solver.parameters.max_time_in_seconds = time
	solver.parameters.num_search_workers = int(threads)
#	status = solver.Solve(model)
	solution_printer = cp_model.ObjectiveSolutionPrinter()
	status = solver.SolveWithSolutionCallback(model, solution_printer)
	print('Finished Solving')
	if status == cp_model.OPTIMAL:
		print('found optimal')
		print('Total edges = %i' % solver.ObjectiveValue())
    
	if status == cp_model.INFEASIBLE:
		print('Infeasible')	
		exit()
     
	if status == cp_model.FEASIBLE:
		print('found feasible')	
		print('Total edges = %i' % solver.ObjectiveValue())

	if status in (cp_model.OPTIMAL, cp_model.FEASIBLE):
		x_m = {}
		y_m = {}
		for (s,t) in SS:
			y_m[s,t] = solver.Value(y[s,t])
		
		for (g,s) in W:
			x_m[g,s] = solver.Value(x[g,s])
			
		print('saving pickle file')
		mytuple = (x_m,y_m,flist)
		pickle.dump(mytuple, open( "saveX.p", "wb"))

		(gr,mini1,mini2,mini3) = CreateMiniGroups(x_m,G,W,S)
		return mytuple 	
			
			
\end{lstlisting}
\end{tiny}

\section{Timetabling CP-SAT Program} \label{appTime}

\begin{tiny}
\begin{lstlisting}[language=Python]
def timetable(x_m, y_m, flist,threads,timeblock,h_mprev = None):
	print('timetable')
	(S,R,RT,C,P,G,L,D,T,Lunch,sd,rd,pd,gd,catalog,divSizeDict,divDict) = flist
	Sorig = list(S)

		
	SallLab = [s for s in Sorig if sd[s].isExtended if sd[s].periods > 1.5]	
	SallLab.append(0)

	Sreg = [s for s in Sorig if s not in SallLab]
			
	if Lunch != -1:
		T = [t for t in T if t != Lunch ]
		
	Y = set([(s,d,t,r) for s in S for d in D for t in T for r in R if (sd[s].roomtype == rd[r].roomtype) ])
	W = set([(g,s) for g in G for s in S if (sd[s].course_num in gd[g].course_num_list) ]  )	
	Q = [(p,s) for p in P for s in S if sd[s].course_num in  pd[p].secNums ]
	
	S3 = [s for s in S if sd[s].isExtended and sd[s].periods == 3]
	
	if Lunch != -1:
		T3 = [t for t in T if ((t+2 < Lunch) or ((t > Lunch) and t+2 <= max(T) )) ]
	else:
		T3 = [t for t in T if t+2 <= max(T)  ]
		
	Y3 = set([(s,d,t,r) for s in S3 for d in D for t in T3 for r in R if (sd[s].roomtype == rd[r].roomtype) ])
	Y3h = set([(s,d,t) for s in S3 for d in D for t in T3 ])
	
	S4 = [s for s in S if sd[s].isExtended and sd[s].periods == 4]
	
	if Lunch != -1:
		T4 = [t for t in T if t+3 < Lunch ]
	else:
		T4 = [t for t in T if t+3 <= max(T) ]
		
	Y4 = set([(s,d,t,r) for s in S4 for d in D for t in T4 for r in R if (sd[s].roomtype == rd[r].roomtype) ])
	Y4h = set([(s,d,t) for s in S4 for d in D for t in T4 ])
	
	S2 = [s for s in S if sd[s].isExtended and sd[s].periods == 2]
	if Lunch != -1:
		T2 = [t for t in T if ((t+1 < Lunch) or ((t > Lunch) and t+1 <= max(T) )) ]
	else:
		T2 = [t for t in T if t+1 <= max(T) ]
		
	Y2 = set([(s,d,t,r) for s in S2 for d in D for t in T2 for r in R if (sd[s].roomtype == rd[r].roomtype) ])
	Y2h = set([(s,d,t) for s in S2 for d in D for t in T2 ])
	
	PS = [(p,s) for p in P for s in S if s in pd[p].secNums]
	PDT = [(p,d,t) for p in P for d in D for t in T]
	GDT = [(g,d,t) for g in G for d in D for t in T]
	
	PDTS = set([(p,d,t,s) for p in P for d in D for t in T for s in S if (p,s) in Q])
	GDTS = set([(g,d,t,s) for g in G for d in D for t in T for s in S if ( ( (g,s) in W)  )  ])
	GC = [(g,c) for g in G for c in C if c in gd[g].course_num_list ]	
	DTR = set([(d,t,r) for d in D for t in T for r in R])
	print('model started')
	m = cp_model.CpModel()
	SDT = [(s,d,t) for s in S for d in D for t in T]
	
	numRoomInType = {}
	for r in R:
		rt = rd[r].roomtype
		if rt in numRoomInType:
			numRoomInType[rt]+=1
		else:
			numRoomInType[rt]=1
	
	RT = list(numRoomInType.keys())			
	
	
	#################Testing
	SS = [(s,t) for s in S for t in S if s < t if y_m[s,t] > 0.5]
	print('Num sections ',len(S))
	print('Num edges ', len(SS))

	solver = cp_model.CpSolver()
#	solver.parameters.log_search_progress = True
	solver.parameters.num_search_workers = int(threads)
	solver.parameters.max_time_in_seconds = timeblock
#	status = solver.Solve(m)
	solution_printer = cp_model.ObjectiveSolutionPrinter()
	m = cp_model.CpModel()


	h = {(s,d,t): m.NewBoolVar('h%i%i%i' % (s,d,t) ) for (s,d,t) in SDT}

	if h_mprev != None:
		for (s,d,t) in SDT:
			m.AddHint(h[s,d,t],h_mprev[s,d,t])
	

	for s in S:
			m.Add(sum(h[s,d,t] for d in D for t in T if (s,d,t) in SDT ) == sd[s].periods)



	fe4h = {(s,d,t): m.NewBoolVar('fe4h%i%i%i' % (s,d,t) ) for (s,d,t) in Y4h}
	for s in S4:
		m.Add(sum( fe4h[s,d,t] for d in D for t in T if (s,d,t) in Y4h)  == 1 )


	for (s,d,t) in Y4h:
		m.Add(h[s,d,t] + h[s,d,t+1] + h[s,d,t+2] + h[s,d,t+3] >= 4*fe4h[s,d,t] )
	
	for (s,d,t) in Y4h:	
		m.AddImplication(fe4h[s,d,t],h[s,d,t] )
		m.AddImplication(fe4h[s,d,t],h[s,d,t+1] )
		m.AddImplication(fe4h[s,d,t],h[s,d,t+2] )
		m.AddImplication(fe4h[s,d,t],h[s,d,t+3] )


	fe3h = {(s,d,t): m.NewBoolVar('fe3h%i%i%i' % (s,d,t)) for (s,d,t) in Y3h } 

	for s in S3:
		m.Add(sum( fe3h[s,d,t] for d in D for t in T if (s,d,t) in Y3h)  == 1 )

	for (s,d,t) in Y3h:	
		m.AddImplication(fe3h[s,d,t],h[s,d,t] )
		m.AddImplication(fe3h[s,d,t],h[s,d,t+1] )
		m.AddImplication(fe3h[s,d,t],h[s,d,t+2] )

	fe2h = {(s,d,t): m.NewBoolVar('fe2h%i%i%i' % (s,d,t)) for (s,d,t) in Y2h }

	for s in S2:
		m.Add(sum( fe2h[s,d,t] for d in D for t in T if (s,d,t) in Y2h)  == 1 )

	for (s,d,t) in Y2h:	
		m.AddImplication(fe2h[s,d,t],h[s,d,t] )
		m.AddImplication(fe2h[s,d,t],h[s,d,t+1] )
	
	t2times = {(s,d):m.NewBoolVar('t2times%i%i' % (s,d) ) for s in S for d in D }
	
	for s in S: 
		for d in D:
			if sd[s].periods == 1 or not sd[s].isExtended:
				m.Add( sum( h[s,d,t] for t in T  if (s,d,t) in SDT ) <= 1 + t2times[s,d])	


	troomh = {(d,t,rt):m.NewBoolVar('troomh%i%i%s' % (d,t,rt) ) for d in D for t in T for rt in RT}
	for d in D:
		for t in T:
			for rt in RT:
				m.Add(sum( h[s,d,t] for s in S if sd[s].roomtype == rt) <= numRoomInType[rt] + 2*troomh[d,t,rt]) 

	SS = [(s,t) for s in S for t in S if s < t if y_m[s,t] > 0.5]
	tconf = {(s1,s2,d,t):m.NewBoolVar('tconf%i%i%i%i' % (s1,s2,d,t) ) for (s1,s2) in SS for d in D for t in T }
	
	print('3')
	pof = {(p,d): m.NewBoolVar('pof%i%i' % (p,d)) for p in P for d in D}

	for p in P:
		m.Add( sum(pof[p,d] for d in D) <= len(D)-1 )
#		m.Add( pof[p,random.choice(D)] == 0 )
		m.Add( pof[p,p%5] == 0 )
	


	tpof = {p: m.NewBoolVar('tpof%i' % p) for p in P}
	for (p,s) in PS:
		for d in D:
			for t in T:
				m.Add(h[s,d,t] <= tpof[p]).OnlyEnforceIf(pof[p,d].Not())
	comw = {s: 1 for s in S}
	comw[0] = 10
	
	m.Minimize( 10*sum(t2times[s,d] for s in S for d in D)  + sum(tpof[p] for p in P ) + sum(1000*comw[s1]*tconf[s1,s2,d,t]  for (s1,s2) in SS for d in D for t in T) + 100*sum(troomh[d,t,rt] for d in D for t in T for rt in RT) )
	
	startI=0
	if h_mprev != None:
		startI = -1	

	tabu1 = None
	for i in range(startI,2):
		if i == 0:
			limitTo = SallLab
			solver.parameters.max_time_in_seconds = timeblock
			
			
		if i == 1:
			limitTo = Sorig	
			solver.parameters.max_time_in_seconds = timeblock
		for (s1,s2) in SS:
			for d in D:
				for t in T:
					if s1 in limitTo and s2 in limitTo:
						m.Add( h[s1,d,t] + h[s2,d,t] <= 1 + tconf[s1,s2,d,t] )
						
		print('Calling Solver')
		status = solver.SolveWithSolutionCallback(m, solution_printer)
		if status == cp_model.INFEASIBLE:
			print('Infeasible value of i is ',i)
			return None
		if status in (cp_model.OPTIMAL, cp_model.FEASIBLE):
			print('finished sections and common hour')
			Temp104 =  [(s1,s2,d,t) for (s1,s2) in SS for d in D for t in T if solver.Value(tconf[s1,s2,d,t]) > 0.5]
			if len(Temp104) > 0:
				tabu1 = []
				for (s1,s2,d,t) in Temp104:
					print('Conflict: ',s1, s2, sd[s1].course_name, sd[s2].course_name, ' time date ',d,t)
					#SS = [(s,t) for s in S for t in S if s < t if y_m[s,t] > 0.5]
					Temp105 = [g for g in G if (g,s1) in W if (g,s2) in W if x_m[g,s1] > 0.5 if x_m[g,s2] > 0.5]
					for g in Temp105:
						tabu1.append((g,s1))
						tabu1.append((g,s2))
					
					
			U1Ah = [(s,d,t) for (s,d,t) in SDT if s in limitTo if solver.Value(h[s,d,t]) > 0.5]
			U1Bh = [(s,d,t) for (s,d,t) in SDT if s in limitTo if solver.Value(h[s,d,t]) < 0.5]
			
			for q in U1Ah:
				m.AddHint(h[q],1)
			for q in U1Bh:
				m.AddHint(h[q],0)
			
			if i==1:	
				for q in U1Ah:
					m.Add(h[q]==1)
				for q in U1Bh:
					m.Add(h[q]==0)		
	

	if status in (cp_model.OPTIMAL, cp_model.FEASIBLE):
		z = {(s,d,t,r): m.NewBoolVar('z%i%i%i%i' % (s,d,t,r) ) for (s,d,t,r) in Y}
		for (s,d,t) in SDT:					
			m.Add( h[s,d,t] == sum(z[s,d,t,r] for r in R if (s,d,t,r) in Y)  )
		for s in S:
			m.Add(sum(z[s,d,t,r] for d in D for t in T for r in R if (s,d,t,r) in Y ) == sd[s].periods)
		
		fe4 = {(s,d,t,r): m.NewBoolVar('fe4%i%i%i%i' % (s,d,t,r) ) for (s,d,t,r) in Y4}
		for s in S4:
			m.Add(sum( fe4[s,d,t,r] for (d,t,r) in DTR if (s,d,t,r) in Y4)  == 1 )
		
		for (s,d,t,r) in Y4:
			m.Add(z[s,d,t,r] + z[s,d,t+1,r] + z[s,d,t+2,r] + z[s,d,t+3,r] >= 4*fe4[s,d,t,r] )
		
		for (s,d,t,r) in Y4:	
			m.AddImplication(fe4[s,d,t,r],z[s,d,t,r] )
			m.AddImplication(fe4[s,d,t,r],z[s,d,t+1,r] )
			m.AddImplication(fe4[s,d,t,r],z[s,d,t+2,r] )
			m.AddImplication(fe4[s,d,t,r],z[s,d,t+3,r] )
		
		
		fe3 = {(s,d,t,r): m.NewBoolVar('fe3%i%i%i%i' % (s,d,t,r)) for (s,d,t,r) in Y3 } 
		
		for s in S3:
			m.Add(sum( fe3[s,d,t,r] for (d,t,r) in DTR if (s,d,t,r) in Y3)  == 1 )

		for (s,d,t,r) in Y3:	
			m.AddImplication(fe3[s,d,t,r],z[s,d,t,r] )
			m.AddImplication(fe3[s,d,t,r],z[s,d,t+1,r] )
			m.AddImplication(fe3[s,d,t,r],z[s,d,t+2,r] )
		
		fe2 = {(s,d,t,r): m.NewBoolVar('fe2%i%i%i%i' % (s,d,t,r)) for (s,d,t,r) in Y2 }

	
		for s in S2:
			m.Add(sum( fe2[s,d,t,r] for (d,t,r) in DTR if (s,d,t,r) in Y2)  == 1 )

		for (s,d,t,r) in Y2:	
			m.AddImplication(fe2[s,d,t,r],z[s,d,t,r] )
			m.AddImplication(fe2[s,d,t,r],z[s,d,t+1,r] )
		
		
		troom = {(d,t,r):m.NewBoolVar('troom%i%i%i' % (d,t,r) ) for d in D for t in T for r in R}
		for d in D:
			for t in T:
				for r in R:
					m.Add(sum( z[s,d,t,r] for s in S if (s,d,t,r) in Y ) <= 1 + troom[d,t,r])
		
		
		m.Minimize( 10*sum(t2times[s,d] for s in S for d in D) + 5*sum(tpof[p] for p in P ) + sum(1000*tconf[s1,s2,d,t]  for (s1,s2) in SS for d in D for t in T) + 100*sum(troom[d,t,r] for d in D for t in T for r in R) )
		
		print('Calling Solver')
		status = solver.SolveWithSolutionCallback(m, solution_printer)
		if status == cp_model.INFEASIBLE:
			print('Infeasible Rooms')
			return
		if status in (cp_model.OPTIMAL, cp_model.FEASIBLE):
			print('reading z')
		
			print('reading z')
			z_m = {}
			for q in Y:
				z_m[q] = solver.Value(z[q])
	# 			if z_m[q] > 0.5:
	# 				print(q,z_m[q])
			print('computing w')

			h_m = {(s,d,t): solver.Value(h[s,d,t]) for s in S for d in D for t in T }
			w_m = {}
			for (p,d,t) in PDT:
				w_m[p,d,t] = sum([h_m[s,d,t] for s in S if (p,s) in PS ] )
	
	
			print('computing u')
			u_m = {(g,d,t,s) : 0 for (g,d,t,s) in GDTS}


			for (g,s) in W:
				if x_m[g,s] > 0.5:
					for d in D:
						for t in T:
							u_m[g,d,t,s] = sum( [ z_m[s,d,t,r] for r in R if (s,d,t,r) in Y ] ) 
		
	
		
			savedSol = (flist,x_m,z_m,w_m,u_m,GDTS)
			pickle.dump(savedSol, open("solX.p","wb") ) 
			return (h_m,tabu1,solver.ObjectiveValue())
	
	# 			
		else:
			print('Did not solve')		



\end{lstlisting}
\end{tiny}

\end{appendices}	

\bibliography{sec}
\bibliographystyle{amsalpha}
   
\vspace{5mm}

\noindent
Joshua S. Friedman \\
Department of Mathematics and Science \\
\textsc{United States Merchant Marine Academy} \\
300 Steamboat Road \\
Kings Point, NY 11024 \\
U.S.A. \\
e-mail: FriedmanJ@usmma.edu, CrownEagle@gmail.com

\noindent
Comments and suggestions are welcome 

\end{document}